\documentclass[lettersize,journal]{IEEEtran}
\usepackage{amsmath,amsfonts}
\usepackage{algorithm}
\usepackage{array}
\usepackage[caption=false,font=normalsize,labelfont=sf,textfont=sf]{subfig}
\usepackage{textcomp}
\usepackage{stfloats}
\usepackage{url}
\usepackage{verbatim}
\usepackage{graphicx}
\usepackage{cite}
\usepackage{bm}
\usepackage{enumitem}
\usepackage[noend]{algpseudocode}
\usepackage{color}
\usepackage{pifont}
\usepackage{wrapfig,lipsum,booktabs}
\algnewcommand\algorithmicinput{\textbf{Input:}}
\algnewcommand\Input{\item[\algorithmicinput]}
\algnewcommand\algorithmicoutput{\textbf{Output:}}
\algnewcommand\Output{\item[\algorithmicoutput]}

\usepackage[svgnames,table,xcdraw]{xcolor}
\definecolor{customred}{HTML}{C25759}
\definecolor{customblue}{HTML}{06007A}
\definecolor{customdarkred}{HTML}{7B4435}
\definecolor{customdarkgreen}{HTML}{447059}
\definecolor{customdarkyellow}{HTML}{CEA33F}
\definecolor{customedarkgray}{HTML}{595959}

\definecolor{customehighlight}{HTML}{ffffff}

\usepackage[most]{tcolorbox}

\newtcbox{\redbox}{on line,
    boxsep=2.5pt,
    boxrule=0pt,
    left=-0.5pt,right=-0.5pt,top=-0.9pt,bottom=-1pt,
    colback=customred, 
    colframe=customred, 
    colupper=white, 
    fontupper=\small,
    arc=0.5mm
}
\newtcbox{\metricone}{on line,
    boxsep=2.5pt,
    boxrule=0pt,
    left=-0.9pt,right=-0.9pt,top=-0.7pt,bottom=-0.9pt,
    colback=customdarkred, 
    colframe=customdarkred, 
    colupper=white, 
    fontupper=\small,
    arc=0.5mm
}
\newtcbox{\metrictwo}{on line,
    boxsep=2.5pt,
    boxrule=0pt,
    left=-0.9pt,right=-0.9pt,top=-0.7pt,bottom=-0.9pt,
    colback=customdarkgreen, 
    colframe=customdarkgreen, 
    colupper=white, 
    fontupper=\small,
    arc=0.5mm
}
\newtcbox{\metricthree}{on line,
    boxsep=2.5pt,
    boxrule=0pt,
    left=-0.9pt,right=-0.9pt,top=-0.7pt,bottom=-0.9pt,
    colback=customdarkyellow, 
    colframe=customdarkyellow, 
    colupper=white, 
    fontupper=\small,
    arc=0.5mm
}

\newcommand{\vcircle}[1]{%
    \raisebox{-0.3ex}{\Large\textcolor{gray}{\ding{108}}}%
    \llap{\raisebox{0.1ex}{\makebox[0pt][c]{\small\textbf{\textcolor{white}{#1}}}}\hspace{4.6pt}}%
}

\newcommand{\systemname}{\emph{VISTA}}
\newcommand{\viewA}{\textit{Label View}}
\newcommand{\viewB}{\textit{Issue Pattern View}}
\newcommand{\viewC}{\textit{Embedding View}}
\newcommand{\viewD}{\textit{Sample View}}
\newcommand{\viewE}{\textit{Annotation Panel}}

\usepackage{soul}
\sethlcolor{customehighlight}

\usepackage{multirow}
\usepackage[colorlinks=true,linkcolor=customblue,citecolor=customblue,urlcolor=black]{hyperref}

\let\titleold\title
\renewcommand{\title}[1]{%
    \titleold{#1}
    \gdef\thetitle{#1}
}

\begin{document}

\title{\emph{VISTA}: A Visual Analytics Framework to\\Enhance Foundation Model-Generated Data Labels}

\author{Xiwei Xuan$^{1, \dag}$, Xiaoqi Wang$^{2}$, Wenbin He$^2$, Jorge Piazentin Ono$^2$, Liang Gou$^{3, \dag}$, Kwan-Liu Ma$^1$, and Liu Ren$^2$
\thanks{$^1$ Department of Computer Science, University of California, Davis, CA, USA. E-mails: \{xwxuan, klma\}@ucdavis.edu.}
\thanks{$^2$ Bosch Center for Artificial Intelligence (BCAI), Bosch Research North America. E-mails: \{xiaoqi.wang, wenbin.he2, jorge.piazentinono, liu.ren\}@us.bosch.com.}
\thanks{$^3$ Splunk Technology, San Jose, CA, USA. E-mail: lgou.psu@gmail.com.}
\thanks{$\dag$ This work was done when the authors worked with Bosch Research North America.}
}

\markboth{Journal of \LaTeX\ Class Files,~Vol.~14, No.~8, August~2021}%
{Shell \MakeLowercase{\textit{et al.}}: A Sample Article Using IEEEtran.cls for IEEE Journals}

\maketitle

\IEEEpubid{0000--0000/00\$00.00~\copyright~2021 IEEE}

\begin{abstract}
The advances in multi-modal foundation models (FMs) (e.g., CLIP and LLaVA) have facilitated the auto-labeling of large-scale datasets, enhancing model performance in challenging downstream tasks such as open-vocabulary object detection and segmentation. However, the quality of FM-generated labels is less studied as existing approaches focus more on data quantity over quality. This is because validating large volumes of data without ground truth presents a considerable challenge in practice. Existing methods typically rely on limited metrics to identify problematic data, lacking a comprehensive perspective, or apply human validation to only a small data fraction, failing to address the full spectrum of potential issues. To overcome these challenges, we introduce \systemname{}, a visual analytics framework that improves data quality to enhance the performance of multi-modal models. Targeting the complex and demanding domain of open-vocabulary image segmentation, \systemname{} integrates multi-phased data validation strategies with human expertise, enabling humans to identify, understand, and correct hidden issues within FM-generated labels. Through detailed use cases on two benchmark datasets and expert reviews, we demonstrate \systemname{}'s effectiveness from both quantitative and qualitative perspectives.

\end{abstract}


\IEEEpubidadjcol

\section{Introduction} 
\label{sec:intro}
Recent advancements have equipped foundation models (FMs) with exceptional capabilities, enabling them to interpret complex language or visual environments~\cite{kirillov2023segment,radford2021learning}. Specifically, FMs can autonomously generate textual descriptions for extensive visual data, marking a prevailing trend in recent techniques---integrating FMs in data pipelines to enhance the volume and diversity of training data~\cite{wang2024use,peng2023kosmos}. This incorporation significantly contributes to various vision-language models, facilitating tasks like image captioning, which generates descriptive sentences for images~\cite{liu2024visual}; phrase grounding, connecting phrases to image regions~\cite{peng2023kosmos}; and open-vocabulary image segmentation (OVIS), assigning labels from an expansive vocabulary to image segments~\cite{wang2024use}. In their approaches, multi-modality FMs like SAM~\cite{kirillov2023segment}, LLaVA~\cite{liu2024visual}, or CLIP~\cite{radford2021learning} are leveraged to produce image-text data in grand scales.
Existing approaches mainly focus on expanding data volumes~\cite{peng2023kosmos,liu2024visual}, which improve model performance to an extent. However, such an emphasis on quantity overshadows a critical aspect of data quality, leading to decreased performance or unexpected model behaviors~\cite{dai2023emu}. Aggregating problematic data in real-world applications can amplify hidden issues, posing significant risks in complex and unpredictable environments. 

In this regard, recent studies~\cite{dai2023emu,touvron2023llama} have shown that models trained on a small, high-quality subset of data significantly outperform those trained on larger datasets of unverified quality.
However, the demanding task of multi-modal data quality enhancement is challenging. Regardless of the complexity caused by large volumes and multi-modality, data from FMs are often generated with self-governance and lack verified ground truth. Efforts to enhance data quality typically involve two categories: (1) using model metrics or data statistics to eliminate problematic data~\cite{schuhmann2021laion,radford2021learning,xu2023demystifying,dai2023emu}; and (2) leveraging humans for issue correction~\cite{kirillov2023segment,touvron2023llama}. Nevertheless, existing automatic methods tend to be single- or narrow-faceted, and human-assisted approaches are often only conducted on a small subset due to the considerable cost of validating prohibitive data volumes. Since data quality should indeed be evaluated from multiple perspectives and quality enhancement should cover data inclusively, existing methods are insufficient to meet the desired objectives~\cite{touvron2023llama,kirillov2023segment,schuhmann2021laion,xu2023demystifying}. Thus, there is a pressing need in this field to meet the increasing requirements of high-quality datasets on large scales.

To address the aforementioned challenges, we introduce \systemname{}, a visual analytics framework to enhance the quality of FM-generated data. We focus on OVIS, one of the most typical yet challenging problems in multi-modality, to analyze, deconstruct, and address this complex challenge.
Considering the critical quality issues raised by the OVIS data pipeline, we sort through different error types with multifaced data quality metrics, summarize them in the visual interface, and design interactive workflow for efficient issue detection, validation, and mitigation with experts in the loop.
To the best of our knowledge, \systemname{} stands as a pioneering framework that addresses this challenge by integrating the strengths of automated computation and human expertise. 
Specifically, this work makes several key contributions by:
\begin{enumerate}[label=\textbullet, align=left, leftmargin=0pt, labelindent=!, listparindent=\parindent, labelwidth=0pt, itemindent=!, itemsep=1pt, parsep=1pt]
\item introducing an automatic issue detection and summarization approach that systematically organizes multi-modal data, quantifying data quality from various perspectives, and summarizing potential issues at scale, which offers a comprehensive overview for efficient problem-solving.
\item designing an interactive visual interface that seamlessly in-
\newpage
\noindent tegrates with algorithmic methods to form a novel visual analytics framework, \systemname{}. which enhances the quality of FM-generated multi-modal data with effective human assistance.
\item demonstrating the effectiveness of \systemname{} through use cases and an expert study, which assesses quantitative performance gains it delivers to OVIS and qualitative enhancements in helping humans detect, understand, and resolve data issues.
\end{enumerate}
We hope this work can stimulate further considerations for quality-centric technologies in the development of next-generation machine learning (ML) systems.

\section{Related Work}
\label{sec:related_work}
\subsection{Foundation Model-Involved Data Pipeline.}
\label{rw:1}
FMs' capability to understand the diverse multi-modal world facilitates their applications in auto-labeling, which automatically generates textual descriptions of images/videos~\cite{peng2023kosmos,liu2024visual}. This has spurred many efforts to leverage FMs in data pipelines, such as general-purpose image-language understanding~\cite{liu2024visual}, referring between noun phrases and bounding boxes~\cite{peng2023kosmos}, open-vocabulary image segmentation~\cite{wang2024use}. While extending data volumes has contributed to performance improvement, there are increased quality concerns. For instance, Dai et al.~\cite{dai2023emu} prove that improving the quality of only a small data amount significantly boosts image generation performance. Touvron et al.~\cite{touvron2023llama} demonstrates that high-quality data acquired by supervised fine-tuning (SFT) can notably improve model performance. However, the efficiency of data quality enhancement remains a persistent challenge---Data validated by humans has verified quality with considerable costs, thus existing work often opts for validating a sampled subset~\cite{touvron2023llama,kirillov2023segment}.
On the other hand, automatic methods utilize quality metrics to filter out unsatisfactory data~\cite{schuhmann2021laion,xu2023demystifying}, which may miss the big picture and overlook many crucial issues~\cite{yang2024foundation}. \systemname{} aims to fill the existing gap with effective human-AI collaboration---bypassing the disadvantages including high manual labeling costs and limited capability of static filtering.

\vspace{-10pt}
\subsection{Human-in-the-Loop Data Labeling and Validation}
\label{rw:2}
Visual analytics (VA) plays a crucial role in labeling and validating expansive datasets~\cite{bernard2017comparing,chegini2019interactive,khayat2019vassl,xuan2022vac}, providing data exploration capabilities to expedite the costly annotation process~\cite{zhang2019cost,wang2024simulation}. VA tools for label creation develop visualization techniques \cite{choi2019aila,zhang2022onelabeler,bernard2017unified,moehrmann2011improving} or automated algorithms \cite{desmond2021increasing,chegini2020interactive} to facilitate label collection, which still demands substantial human involvement or fail to handle more complex multi-modal data. Label or model validation systems assume labels have already been produced and focus on validating their quality~\cite{hoferlin2012inter,chen2020oodanalyzer,chen2021towards,yang2022diagnosing,vajiac2022trafficvis,chen2021interactive,yang2023interactive,chen2024enhancing,sharifi2023perspective,bauerle2020classifier,xuan2024attributionscanner,xuan2024suny}. Techniques such as dimensionality reduction and uncertainty estimation have been employed to verify crowd-sourced image datasets and detect problematic samples \cite{liu2018interactive,xiang2019interactive,zhang2023labelvizier,xuan2024slim}, which can be resource-intensive. Hoque et al.~\cite{hoque2022visual} introduced Visual Concept Programming that uses semantic segmentation models and allows experts to infuse domain knowledge, scaling up data labeling through automated functions. Therefore, facing multi-modal data that are often more complex due to varied modalities, heterogeneous in structure, and massive in volume, we introduce \systemname{} to efficiently address data quality concerns with humans in the loop.

\setlength{\floatsep}{2pt plus 2pt minus 2pt}
\setlength{\textfloatsep}{2pt plus 2pt minus 2pt}
\setlength{\intextsep}{2pt plus 2pt minus 2pt}
\begin{figure}[th!]
\setlength{\abovecaptionskip}{0pt}
  \centering\includegraphics[width=\linewidth]{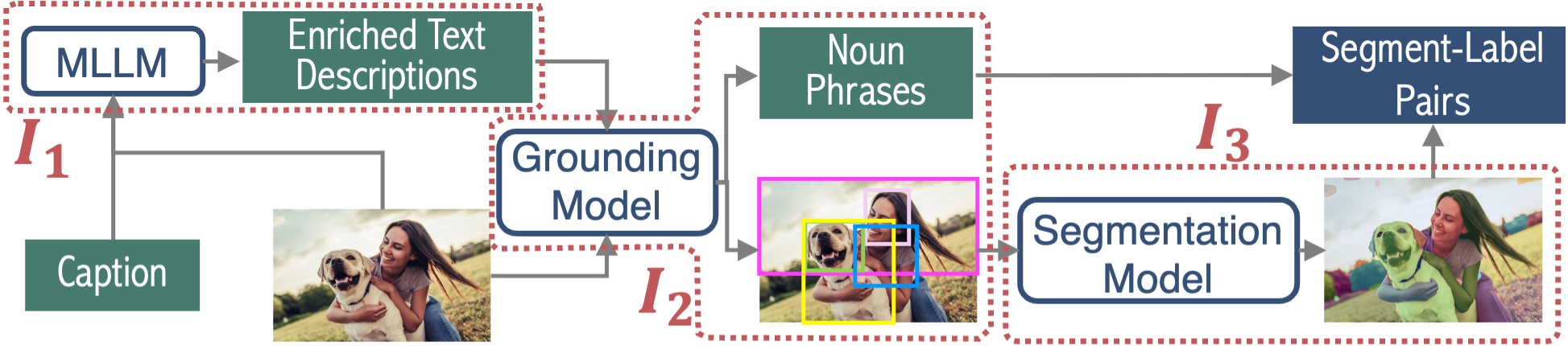}
  \caption{
  OVIS data generation pipeline, involving FMs to construct segment-label pairs. {\color{customred}Red outlines} indicate potential stages with data quality concerns.
  }
  \label{fig:data_pipeline}
\end{figure}

\vspace{-10pt}
\subsection{Open-Vocabulary Image Segmentation}
\label{rw:3}
OVIS marks a significant evolution in autolabeling, enabling models to recognize fine-grained semantics in the dynamic real world. 
Two primary approaches define the current landscape of this domain.
Training-based approaches~\cite{huynh2022open,xu2022simple,xu2022groupvit,xu2023open,xu2023side,zou2023generalized} use a web-scale dataset to train or fine-tune a sophisticated model, which can produce good results but are computationally intensive. The performance of these approaches is affected by multiple factors including model architectures, training data, loss functions, and optimization strategies~\cite{zou2023generalized,xu2022groupvit,xu2023open,xu2023side}. On the other hand, training-free OVIS either explores the attention mechanisms of FMs~\cite{zhou2022extract,lan2024proxyclip,shao2024explore} or constructs a reference set to perform semantic segmentation via information retrieval~\cite{shin2022reco,karazija2024,barsellotti2024fossil,gui2024knnclip,wang2024image}, while retrieval-based methods provide a valuable evaluation for our framework, whose performance is mainly determined by the quality of reference sets.

\section{Motivation and \systemname{} Framework}
\label{sec:problem_and_design}
In this section, we overview the OVIS data pipeline with FMs, discuss challenges and requirements for quality enhancement, and introduce \systemname{} framework. Note that \systemname{} is developed in close collaboration with two ML experts working in the industry, who are co-authors of this paper.

\vspace{-10pt}
\subsection{OVIS and Data Pipeline with FMs}
Fig.~\ref{fig:data_pipeline} shows a typical data generation pipeline of OVIS.
A multi-modal large language model (MLLM) processes the input image to produce enriched text descriptions. A grounding model then links noun phrases from descriptions to image regions using bounding boxes, according to which a segmentation model crafts segment masks.

\noindent\textbf{Data quality considerations.} \label{subsubsec:quality_concern}
In Fig.~\ref{fig:data_pipeline}, three {\color{customred} outlined} stages can potentially introduce quality issues, as detailed below.

\begin{enumerate}[label={\color{customred} \( \bm{I}\arabic* \).}, ref={\color{customred} \( \bm{I}\arabic* \)}, align=left, leftmargin=0pt, labelindent=!, listparindent=\parindent, labelwidth=0pt, itemindent=!, itemsep=1pt, parsep=1pt]
\item \textbf{Label issues.}\label{issue:label}
The label generation process with MLLMs can produce contextually irrelevant labels (Fig.~\ref{fig:data_pipeline} ({\color{customred} \( \bm{I_1} \)})).

\item \textbf{Alignment issues.}\label{issue:misalignment}
Grounding process pairs noun phrases with image regions (Fig.~\ref{fig:data_pipeline} ({\color{customred} \( \bm{I_2} \)})). However, contextual information outside an image region can be mixed in its embeddings, leading to inaccurate alignments.

\item \textbf{Image segmentation issues.}\label{issue:segment}
The image segmentation process (Fig.~\ref{fig:data_pipeline} ({\color{customred} \( \bm{I_3} \)})) may generate low-quality masks, such as those without accurate boundaries or meaningful content.
\end{enumerate}

Our discussions with ML experts indicate that not all issues are equally significant. \ref{issue:segment} have become less frequent due to technology advancements~\cite{kirillov2023segment}, while the other two remain significant contributors to data quality problems and require close attention. Consequently, our research focuses on addressing the most critical challenges
related to \ref{issue:label} and \ref{issue:misalignment}.

\vspace{-10pt}
\subsection{Challenges for Quality Enhancement}
Building upon the identified data quality considerations, we delineate the core challenges that must be addressed.
\begin{figure*}[thb]
\setlength{\abovecaptionskip}{0pt}
  \centering
  \includegraphics[width=.9\linewidth]{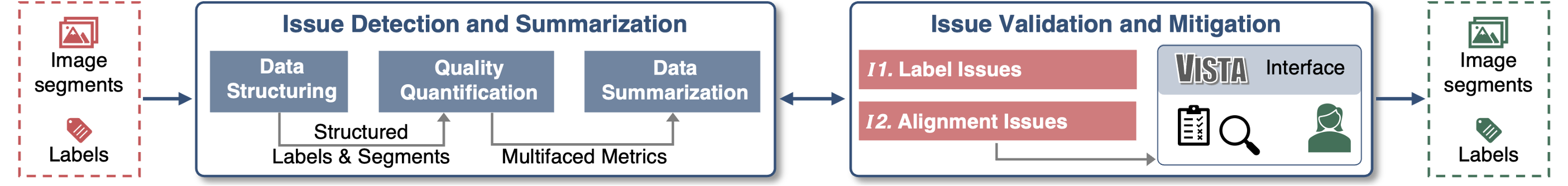}
  \caption{
\systemname{} data quality enhancement framework, comprising the method phase for issue detection and summarization and the visual analytic interface for issue validation and mitigation. With \systemname{}, we enhance foundation model-generated data quality with human assistance.
  }
  \vspace{-10pt}
  \label{fig:voise_framework}
\end{figure*}

\begin{enumerate}[label=\textbf{C\arabic*.}, ref=\textbf{C\arabic*}, align=left, leftmargin=0pt, labelindent=!, listparindent=\parindent, labelwidth=0pt, itemindent=!, itemsep=1pt, parsep=1pt]

\item \textbf{Lack of ground-truth data.} \label{challenge:c1}
As discussed in Sec.~\ref{sec:intro}, FMs' involvement in data pipelines results in data without verifiable ground truth~\cite{schuhmann2021laion,xu2023demystifying,kirillov2023segment}. However, identifying wrong data is foundational for quality validation, and the absence of ground truth makes it challenging to spotlight problematic data, thereby impeding subsequent remediation efforts.

\item \textbf{Complexity of multi-modal information.} \label{challenge:c2}
While we categorize the main data concerns into three types, their interconnected nature prevents them from being viewed in isolation. However, tackling them simultaneously is also not feasible. Such complexity of the multi-modal landscape poses additional challenges, leading to many manual efforts in prior works~\cite{kirillov2023segment,touvron2023llama}, thereby necessitating a nuanced approach to properly disentangling the underlying problems.

\item \textbf{Large volume of noisy data.} \label{challenge:c3}
Significant noise may be present in the large volume of FM-generated data. Sorting through and enhancing them is a formidable task. For instance, considering human has open-world knowledge to validate data quality, the sheer amount of potential errors necessitates substantial manual effort, which is impractical without a strategically developed support system.

\end{enumerate}

\vspace{-10pt}
\subsection{Requirements for Quality Enhancement}
To overcome the identified challenges, combined with the expertise of our collaborated ML experts, we develop the essential requirements that our framework must meet:

\begin{enumerate}[label=\textbf{R\arabic*.}, ref=\textbf{R\arabic*}, align=left, leftmargin=0pt, labelindent=!, listparindent=\parindent, labelwidth=0pt, itemindent=!, itemsep=1pt, parsep=1pt]

\item \textbf{Issue detection from multiple perspectives.} \label{req:r1}
To facilitate human involvement (\ref{challenge:c3}), we need to come up with an algorithmic strategy to discover data issues. Considering the lack of ground truth (\ref{challenge:c1}), analyzing from multiple angles can provide a more comprehensive error coverage. Moreover, the designed metrics should align closely with the data generation (Fig.~\ref{fig:data_pipeline}), where the issues are produced, and disentangle multi-modal data (\ref{challenge:c2}), considering different issue types (\ref{issue:label}, \ref{issue:misalignment}).

\item \textbf{Multimodel alignment analysis at various granularities.} \label{req:r2}
Analyzing image-text alignment is the cornerstone of multi-modal problems, inherently involving various image granularities (\ref{challenge:c2}), i.e., ``\textit{image}---\textit{segment's surrounding region (bounding box)}---\textit{segment}''. Despite introducing noise to visual embeddings, such granularities have positive contributions to our framework by providing additional relationships to analysis, which can serve as additional evaluation perspectives to address our lack of ground truth challenge (\ref{challenge:c1}).

\item \textbf{Systematically organize, explain, and summarize data.} \label{req:r3}
Our framework should systematically organize, explain, and summarize multi-modal data to facilitate validators' understanding (\ref{challenge:c2}, \ref{challenge:c3}). Our framework should identify similar labels, segments, and misalignments, streamlining the detection of different types of issues. In addition, explaining the formed groups is essential to support understanding.

\item \textbf{Usability-driven design to support human involvement.} \label{req:r4}
Human expertise, while invaluable, is finite and can have declining performance, especially when exhausted. Additionally, our target users are ML experts with limited visualization/VA knowledge. Therefore, our interface design should prioritize usability to avoid overwhelming users. Specifically, we should: (1) provide straightforward rather than sophisticated visualizations; (2) design interactive views to present related information simultaneously; (3) present a summary view to enable memory retrieval with a simple glance.
\end{enumerate}

\vspace{-10pt}
\subsection{\systemname{} Framework}

To address the challenges and requirements, we introduce \systemname{}, a framework to enhance FM-generated data quality. An overview of \systemname{} is shown in Fig.~\ref{fig:voise_framework}. Our issue detection and summarization uses statistical or computational strategies (\ref{req:r1}) to deconstruct this complex problem into coherent subgroups with clear summarization (\ref{req:r2}, \ref{req:r3}). Grounded on the summarized issues, the visual analytics system enables human validators to quickly locate, understand, and solve different issues to enhance data quality (\ref{req:r3}, \ref{req:r4}).

\begin{figure*}[bt]
\setlength{\abovecaptionskip}{0pt}
  \centering\includegraphics[width=0.92\linewidth]{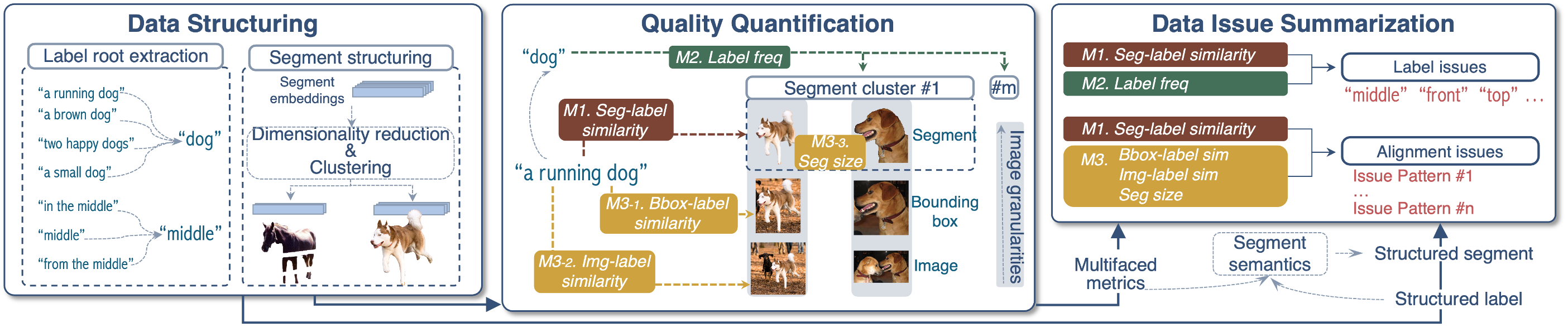}
  \caption{Method in \systemname{} for data issue detection and summarization, with three interlocked phases: label and segment data structuring, quality quantification for different issues, and data issue summarization for forming subgroups with coherent patterns or semantics.}
  \vspace{-10pt}
  \label{fig:backend}
\end{figure*}

\section{Method}
\label{sec:method}
In this section, we discuss the method in \systemname{};s issue detection and summarization backbone, including three phases as shown in Fig.~\ref{fig:backend}. Driven by design requirements, the data structuring phase aims at organizing data from each modality into structured subgroups (\ref{req:r3}). The quality quantification phase focuses on the multiple granularities of data alignments, measuring their quality with multifaced quantification metrics (\ref{req:r1}, \ref{req:r2}). Then, the data issue summarization phase coordinates findings to further structure and explain data with concise semantics or coherent error patterns (\ref{req:r1}, \ref{req:r2}, \ref{req:r3}).

\vspace{-10pt}
\subsection{Data Structuring}
\label{subsec:data_structuring}

\noindent\textbf{Structuring of labels.}
\label{subsubsec:data_structuring_label}
Recent research finds that labels with descriptions encapsulate richer semantics, and thus improve models' open-vocabulary recognition capability~\cite{wu2020phrasecut}. Therefore, noun phrases become the common type of textual label, in the format of \textit{``description} $+$ \textit{\underline{noun}''}. The involvement of FMs further increases their diversity, making it hard for effective exploration. To fulfill \ref{req:r3}, we study this process and find the invariant here is the root of labels---the ``\textit{\underline{noun}}''. For example, for an image with dogs, we can have multiple labels like \textit{``a running \underline{dog}''}, \textit{``pretty \underline{dog}s''}, and \textit{``a brown \underline{dog}''}. Regardless of various descriptions, the root \textit{``\underline{dog}''} is unchanged. Therefore, by extracting the root from noun phrases, we categorize various labels according to their root and form consistent label groups, as depicted in the ``label root extraction'' process in Fig.~\ref{fig:backend}. Specifically, we deploy SpaCy~\cite{spacy} for the root extraction, and record the mapping between each root and their corresponding labels for seamless information retrieval in our analytical framework.

\noindent\textbf{Structuring of segments.}
\label{subsubsec:data_structuring_segment}
Image segments embody a more complex context compared with labels, demanding a meticulous design for organizing them (\ref{req:r3}). As shown in the ``segment structuring'' module in Fig.~\ref{fig:backend}, our method involves a two-step clustering process, comprising the reduction of high-dimensional segment embeddings to a 2D space using UMAP~\cite{mcinnes2018umap} and the clustering with HDBSCAN~\cite{mcinnes2017hdbscan} to identify clusters. The necessity for this intermediate dimensionality reduction arises from the challenges posed by the large amount and dimension of segments. Our experiments with ML experts indicate that directly applying HDBSCAN or other clustering methods on high-dimensional embeddings results in suboptimal clusters. Through results assessment with silhouette scores of different combinations of dimensionality reduction~\cite{van2008visualizing} and clustering~\cite{ester1996density}, this two-step process has emerged as the most effective strategy for our data format and scale, ensuring the creation of meaningful segment groups.

\vspace{-10pt}
\subsection{Quality Quantification}
\label{subsec:quality_quantification}

Let \( I \) denote an image, \( S \) denote an image segment, \( B \) denote the bounding box of the segment, and \( L \) denote a textual label describing the segment, with root \(r\). A multi-modal model, such as \textit{clip\_surgery}~\cite{li2023clip}, including an image encoder and a text encoder, encodes them into embeddings with the same dimension, \( \bm{i} \), \( \bm{s} \), \( \bm{b} \), \( \bm{l} \in \mathbb{R}^{n}\), respectively. Besides, we use \( \frac{\langle x, y \rangle}{\| x \| \cdot \| y \|} \) to represent cosine similarity between \(x\) and \(y\). Now, we delve into our designed data quality metrics.

\begin{enumerate}[label=\metricone{\textit{M1}}, ref=\metricone{\textit{M1}}, align=left, leftmargin=0pt, labelindent=!, listparindent=\parindent, labelwidth=0pt, itemindent=!, itemsep=0pt, topsep=0pt]

\item {\normalsize\sffamily \textit{Segment-label similarity}} \label{metric:seg_label_sim}

\noindent This metric quantifies the alignment between segment and label embeddings, serving as a primary data issue indicator. It is the cosine similarity between \( \bm{s} \) and \( \bm{l} \), represented by:
\begin{equation}\label{eqn:seg_label_sim}\small
sim(\bm{s}, \bm{l}) = \frac{\langle \bm{s}, \bm{l} \rangle}{\| \bm{s} \| \cdot \| \bm{l} \|},
\end{equation}
where a higher \(sim(\bm{s}, \bm{l})\) indicates a stronger correlation.

\end{enumerate}

\begin{enumerate}[label=\metrictwo{\textit{M2}}, ref=\metrictwo{\textit{M2}}, align=left, leftmargin=0pt, labelindent=!, listparindent=\parindent, labelwidth=0pt, itemindent=!, itemsep=0pt, topsep=0pt]
\item {\normalsize\sffamily \textit{Label frequency}} \label{metric:label_freq}

\noindent Building upon the data structuring established in Sec.~\ref{subsec:data_structuring}, we use \(C(r)\) to represent the count of unique segment groups to which a unique root \(r\) is assigned. This metric supports the detection of label issues (\ref{issue:label}), represented by:

\begin{equation}\small
freq(l) = C(root(l)) =  C(r).
\end{equation}
Labels with the same root should not be capable of describing numerous varieties of segments. Therefore, a higher \(freq(l)\) is correlated with ambiguous terms or other potential mistakes.
\end{enumerate}

\begin{enumerate}[label=\metricthree{\textit{M3}}, ref=\metricthree{\textit{M3}}, align=left, leftmargin=0pt, labelindent=!, listparindent=\parindent, labelwidth=0pt, itemindent=!, itemsep=0pt, topsep=0pt]
\item {\normalsize\sffamily \textit{Multi-granularity image-label alignment}} \label{metric:granularity}

\noindent This metric quantifies alignment between image and text (\ref{issue:misalignment}), with a specific focus on image granularities. As discussed in \ref{issue:misalignment}, additional context information outside a segmentation mask can be mixed in segment embeddings, resulting in misalignment where a label is paired with the spatial context of a segment. There are three sub-metrics in this fold:
\begin{equation}
\small
\begin{aligned}
sim(\bm{b}, \bm{l}) = \frac{\langle \bm{b}, \bm{l} \rangle}{\| \bm{b} \| \cdot \| \bm{l} \|};
\end{aligned}
\,
\begin{aligned}
sim(\bm{i}, \bm{l}) = \frac{\langle \bm{i}, \bm{l} \rangle}{\| \bm{i} \| \cdot \| \bm{l} \|};
\end{aligned}
\,
\begin{aligned}
size(S) = \frac{w_S * h_S} {w_I * h_I},
\end{aligned}
\end{equation}
where the first and second are \textit{Bounding box-label similarity} and \textit{Image-label similarity}; the third is \textit{Segment size} that measures the proportion of an image spanned by a segment; \(w_*\) and \(f_*\) represent the width and height in the image space. Specifically, $w_S$ and $h_S$ are defined as the dimensions of the smallest bounding box that can fully enclose the segment $S$.

In the quality quantification phase in Fig.~\ref{fig:backend}, we depict how these metrics assess different perspectives of this pairing relationship in multi-modality (\ref{req:r1}, \ref{req:r2}).
\end{enumerate}

\vspace{-10pt}
\subsection{Data Issue Summarization}
\label{subsec:data_issue_summarization}
In this phase, our goal is to further sort through data, providing concise explanations for coherent groups and identifying consistent issue patterns from our initial findings (\ref{req:r3}), making the information ready to be presented at our visual analytics interface (\ref{req:r4}). Aligned with previous works~\cite{radford2021learning,schuhmann2021laion}, we leverage segment-label similarity (\ref{metric:seg_label_sim}) as the foundation to identify problematic data, as it represents the kernel of this pairing problem. Additionally, with the consideration discussed in \ref{req:r1}, other data issue perspectives are not overlooked.

\subsubsection{Segment semantics discovery}
In Sec.~\ref{subsec:data_structuring}, we have structured segments and labels. However, unlike labels in natural language, the interpretability of image information is not as direct~\cite{wei2023diffusion}.
To offer quick insights to human validators into the key content of each segment group, 
\begin{algorithm} [bht!]
\caption{Discovering semantics for a segment cluster}
\label{alg:seg_semantics}
\begin{algorithmic}[1]
\Input Segment set \(\bm{S}\), set of the corresponding labels \(\bm{L}\), segment and label embeddings $\bm{s}, \bm{l} \in \mathbb{R}^n$, segment 2D embeddings $\bm{c_s} \in \mathbb{R}^2$.
\Output Concise semantics of the segment cluster \(\bm{S}\).
\For{each \(S_i \in \bm{S}\) and \(L_j \in \bm{L}\)}
    \State $sim(\bm{s}_i, \bm{l}_j) = \frac{\langle \bm{s}_i, \bm{l}_j \rangle}{\| \bm{s}_i \| \cdot \| \bm{l}_j \|}$ \textcolor{gray}{\footnotesize// Cosine similarity}
    \State \(kde(\bm{c}_{\bm{s}_i}) = \frac{1}{Nh^2} \sum_{j=1}^{N} K\left( \frac{\bm{c}_{\bm{s}_i} - \bm{c}_{\bm{s}_j}}{h} \right)\) \textcolor{gray}{\footnotesize// KDE}
\EndFor
\State \( r \gets \) [ ] \textcolor{gray}{\footnotesize// Initialize ranking list} 
\For{each segment-label pair \((S_i, L_j) \in (\bm{S, L})\)}
    \State \( score \) \(\gets\) \(kde(\bm{c}_{\bm{s}_i}) \) \(\times\) \(sim(\bm{s}_i, \bm{l}_j)\) \textcolor{gray}{\footnotesize// ranking score}
    \State \(r\).append\(((S_i, L_j, score))\)
\EndFor
\State $count = \lceil p\% \cdot |r| \rceil$ \textcolor{gray}{\footnotesize// top $p\%$}
\State $L_{top} \gets r[argsort(-r.score)[:count]].L$ \textcolor{gray}{\footnotesize// labels in top $p\%$}
\State \Return $L_{top}.root.unique()$ \textcolor{gray}{\footnotesize// unique roots}
\end{algorithmic}
\vspace{-5pt}
\end{algorithm}
we endow them with semantic descriptions through our semantics discovery derived through Alg.~\ref{alg:seg_semantics}.

Given a clustered set of segments \(\bm{S}\), and the set of corresponding labels for each segment \(\bm{L}\). Each segment-label pair has a similarity quantification \ref{metric:seg_label_sim} (line 2 of Alg.~\ref{alg:seg_semantics}). To avoid the impact of clusters' outliers, we focus more on segments densely located within the cluster according to Kernel Density Estimation (KDE), representing the density of neighboring segments within the 2-dimensional embedding space (line 3 of Alg.~\ref{alg:seg_semantics}). The product of the segment-label similarity and the KDE value for each segment yields a ranking score, identifying top similar labels to the central of densely clustered segments, i.e., the cluster's semantics (\ref{req:r3}).

\subsubsection{Label issue summarization}
\label{subsubsec:method_label_issue_sum}

To spotlight the most dominant label issues that happen to groups of labels (i.e., \textit{root} (\ref{req:r4})\textbf{}, we leverage segment-label similarity (\ref{metric:seg_label_sim}) and label frequency (\ref{metric:label_freq}) in this step (\ref{req:r1}). \ref{metric:label_freq} is computed for each label group while \ref{metric:seg_label_sim} is based on a single segment-label pair. To make both describe group-level issues, per the recorded mapping between root and labels (refer to Sec.~\ref{subsubsec:data_structuring_label}), we obtain the median of segment-label similarities corresponding to each root. Therefore, for each label group (root), we have two metrics, representing its problematic level (\ref{metric:seg_label_sim}) and whether they are sketchy or ambiguous to be assigned to a wide variety of segments (\ref{metric:label_freq}), respectively. To provide a holistic analysis of this focal issue, we decided to employ a dual-metric approach in our interface without merging them, which will be discussed in detail in Sec.~\ref{subsec:interface_label_issue}.

\subsubsection{Alignment issue summarization}
\label{subsubsec:method_slice_finding}
The alignment issue (\ref{issue:misalignment}) concurrently elucidated by the segment-label similarity (\ref{metric:seg_label_sim}) and multi-granularity image-label alignment (\ref{metric:granularity}) (\ref{req:r1}). Specifically, \ref{metric:granularity} comprises three sub-metrics for a thorough validation across various granularities (\ref{req:r2}).
Nevertheless, integrating these sub-metrics with the segment-label similarity poses a challenge---How can we distill them into a clear summary, enabling straightforward error pattern identification?

To tackle this challenge, we employ DivExplorer~\cite{pastor2021looking}, a tool designed to find common attributions associated with particular outcomes. For each segment-label pair, we regard \ref{metric:seg_label_sim} as the indicator of misalignment. Consistent with common set-ups~\cite{schuhmann2021laion,xu2023demystifying}, we set its global median as the default value of the threshold---If the value of \ref{metric:seg_label_sim} falls below it, the pair is flagged for misalignment. We then convert the continuous values of each sub-metric in \ref{metric:granularity} into three discrete categories with the same size following common practices---based on their percentiles~\cite{pastor2021looking,zhang2022sliceteller,shetiya2022fairness}, $33\%$, $66\%$ are used in our case. Note that such thresholds are adjustable by users. DivExplorer leverages binary misalignment indicator (outcome) and categorized metrics (attributions) to identify subgroups exhibiting consistent error patterns as per \ref{metric:granularity}. This algorithmic approach offers a solid foundation for human validators to pinpoint and comprehend the primary misalignment patterns (\ref{req:r4}).

\vspace{-10pt}
\subsection{Online Evaluation}\label{subsec:online_eval}
To quantitatively evaluate the capability of \systemname{} in enhancing the quality of datasets with segment-label pairs, we adopt a training-free OVIS method~\cite{tip-adapter} that performs semantic segmentation by retrieving a reference set with segment-label pairs, enabling real-time evaluation of data quality. Specifically, we compare the OVIS performance when using the original and \systemname{}-produced data, respectively.

\begin{figure*}
\setlength{\abovecaptionskip}{0pt}
  \centering
  \includegraphics[width=\linewidth]{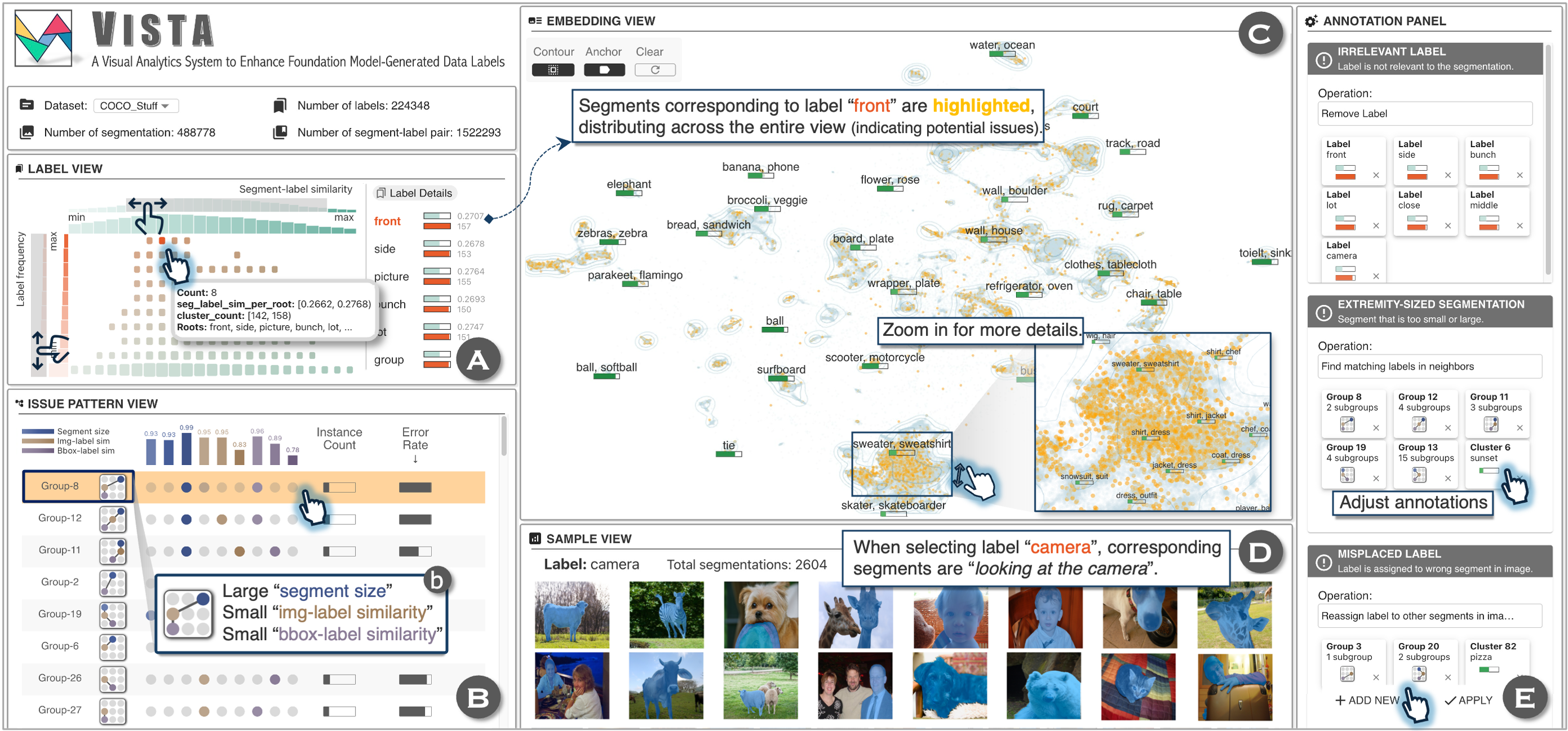}
  \caption{
The system interface of \systemname{} supports the quality enhancement of foundation-model generated labels, integrating: \vcircle{A} \viewA{} presents groups of labels in a grid-based layout, allowing for label issue validation; \vcircle{B} \viewB{} organizes segment-label pairs into data groups based on their common patterns, streamlining the identification and handling of alignment issues; \vcircle{C} \viewC{} is a 2D projection of segment embeddings where users can browse segment clusters or data instances; \vcircle{D} \viewD{} provides a closer look at the selected samples, allowing for in-depth analysis; \vcircle{E} \viewE{} is a utility pane enabling human validators to annotate and correct identified issues, which also provides a summary of the annotation status.
  }
  \vspace{-10pt}
  \label{fig:teaser}
\end{figure*}

Let \((S_{train}, L_{train})\) denote segment-label pairs generated by FMs or enhanced by \systemname{}, with $N$ unique segments. \((S_{test}, L_{test})\) be the ground-truth segment-label pairs in the test set, with $M$ labels. Firstly, we find similar segments and similar labels between $test$ and $train$ using similarity quantifications $sim_{S}$ and $sim_{L}$. We obtain the similarity between \(S_{test}\) and \(S_{train}\) by:

\begin{equation}\label{eqn:eval1}
sim_{S} = softmax((\bm{S}_{test} \cdot \bm{S}_{train}^T) \times \beta_1),
\end{equation}
where \(\bm{S}_{test}\), \(\bm{S}_{train}\) are embeddings of \(S_{test}\), \(S_{train}\); \(sim_{S} \in \mathbb{R}^{1 \times N}\). For all unique labels \(l_{train} \in L_{train}\), we obtain their similarity with \(L_{test}\) labels by:

\begin{equation}\label{eqn:eval2}
sim_{L}^{(i)} = softmax((\frac{\bm{o}^i_{train}}{\|\bm{o}^i_{train}\|}\cdot\bm{L}_{train} \cdot \bm{L}_{test}^T) \times \beta_2), i\in(0,N),
\end{equation}
where \(\bm{o}^i_{train}\) is the binary vector of labels for the i-th segment \(s^i_{train}\), \(\bm{L_{train}}\), \(\bm{L_{test}}\) are embeddings of unique \(L_{train}\), \(L_{test}\), respectively. We have \(sim_{L} \in \mathbb{R}^{N \times M}\). Note that $\beta_1$ and $\beta_2$ are temperatures for softmax, with a lower value considering more neighbors and a higher value leading to the focus on the top similar ones. An ablation study of parameters $\beta_1$ and $\beta_2$ is provided in the supplementary.
Lastly, the predicted labels for \(s_{test}\) is computed by: 
\begin{equation}\label{eqn:eval3}
\hat{l}_{test}={argmax}_{l_{test}} (sim_{S} \times sim_{L}).
\end{equation}
This way, we obtain OVIS results by this referring process. Consequently, we can measure its performance with mIoU and pixel accuracy, the common OVIS validation metrics, by:
\begin{align}
mIoU &= \frac{1}{N}\sum^N_{i=1}\frac{\textit{\text{Area of Overlap}}}{\textit{\text{Area of Union}}} \label{eqn:mIoU} \\
pixel\_acc &= \frac{\textit{\text{\# of Correctly Classified Pixels}}}{\textit{\text{Total \# of Pixels}}} \label{eqn:pixel_acc}
\end{align}

\section{Visual Interface}
\label{sec:interface}


\subsection{Interface Overview}

\systemname{}'s interface, as shown in Fig.~\ref{fig:teaser}, is designed data quality enhancement and consists of five coordinated views:

\begin{enumerate}[align=left, leftmargin=0pt, listparindent=\parindent, labelwidth=0pt, itemindent=0pt, itemsep=0pt, topsep=0pt, label={}]

\item \vcircle{A} \textbf{\viewA{}}: This view facilitates the interactive exploration of summarized label issues (\ref{issue:label}), helping users identify problematic label groups. As shown in Fig.~\ref{fig:teaser} \vcircle{A}, we design a grid visualization showing the 2-dimensional information provided by segment-label similarity (\ref{metric:seg_label_sim}, x-axis) and label frequency (\ref{metric:label_freq}, y-axis), the two key metrics for pinpointing crucial label issues (refer to Sec.~\ref{subsubsec:method_label_issue_sum}) (\ref{req:r1}, \ref{req:r3}). Each rectangular cell represents one or more label groups with size indicating the group count, while as smaller (\ref{metric:seg_label_sim}) and larger (\ref{metric:label_freq}) indicate more troublesome and dominant labels, they are spotlighted by cells closer to the top left of this label grid (\ref{req:r3}, \ref{req:r4}). Validators can brush histograms at each axis to select and check a label group (represented by the root, such as ``front'' in Fig.~\ref{fig:teaser}) to investigate further information. 

\item \vcircle{B} \textbf{\viewB{}}: Presenting summarized alignment issues (\ref{issue:misalignment}), this view allows users to validate predominant error patterns in the alignment between image segment and labels, corresponding to \ref{metric:seg_label_sim} and \ref{metric:granularity} (\ref{req:r2}, \ref{req:r3}). As shown in Fig.~\ref{fig:teaser} \vcircle{B}, we visualize alignment issue summarization (refer to Sec.~\ref{subsubsec:method_slice_finding}) with an UpSet visualization~\cite{lex2014upset}. The discretized sub-metrics of \ref{metric:granularity} are integrated with a matrix layout with nine columns, with the top bars showing the frequency of each attribution. For example, as shown in Fig.~\ref{fig:teaser}, 0.93 of ``small segment size'' means 93\% of samples with ``small segment size'' are flagged as ``wrong labels''. Each row represents one data group with coherent issues, with the corresponding attribution highlighted with colored circles. As \ref{metric:seg_label_sim} is a misalignment (error) indicator, we visualize the calculated ``error rate'' as horizontal bars for each data group.

\begin{figure*}
\setlength{\abovecaptionskip}{0pt}
  \centering
  \includegraphics[width=\linewidth]{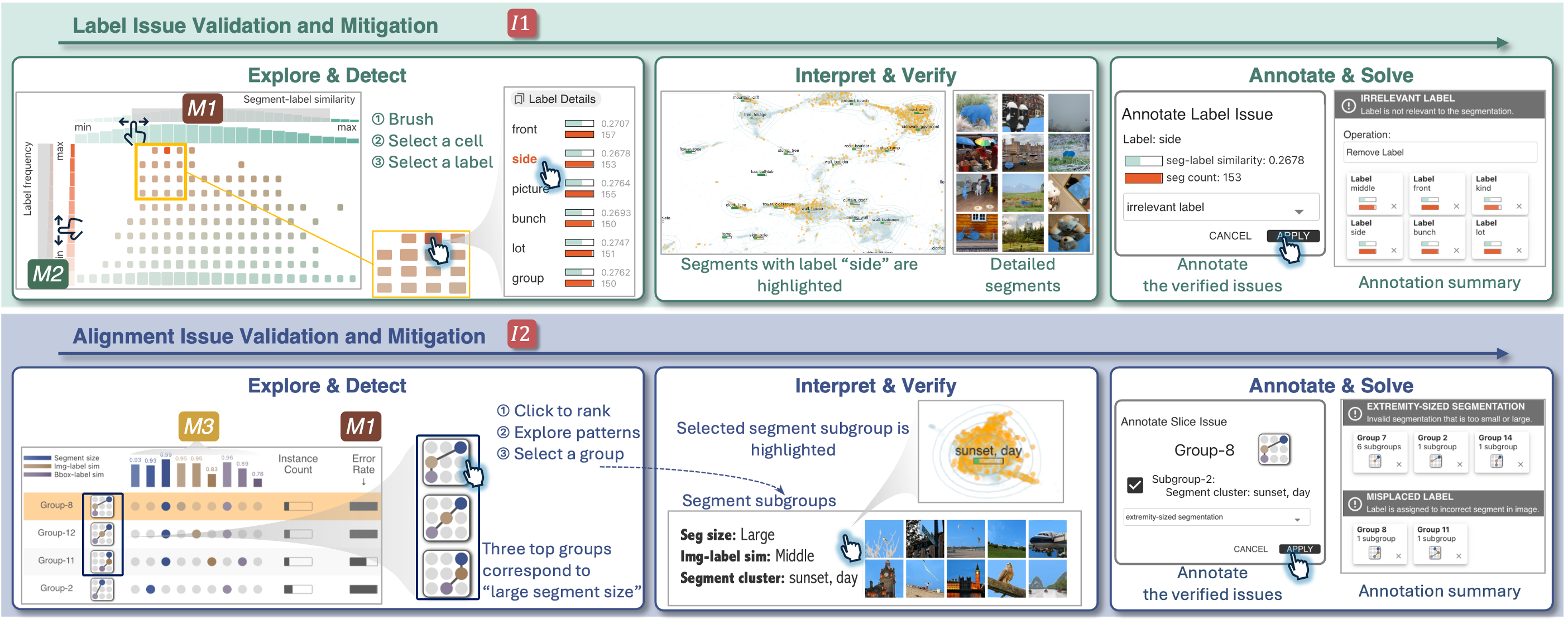}
  \caption{Analytical workflow for human validators to conduct issue validation and mitigation with \systemname{}'s visual interface, with the top and bottom phases addressing \ref{issue:label} and \ref{issue:misalignment}, respectively. Each phase comprises three stages. Humans interact with the \viewA{} or the \viewB{} to \textit{``Explore \& Detect''} issues; \textit{``Interpret \& Verify''} errors by inspecting corresponding segment distributions or data samples through \viewC{} and \viewD{}; and lastly, \textit{``Annotate \& Solve''} verified issue with the \viewE{}, which summarizes annotation records under different categories. 
  }
  \vspace{-10pt}
  \label{fig:analytical_workflow}
\end{figure*}

With the UpSet visualization, we guide validators to explore and detect predominant misalignment patterns by making all columns rankable to prioritize more problematic groups (\ref{req:r4}). In addition, we further observe that the matrix layout is insufficient for quick pattern identification---users may need interactions to confirm related attributions. To facilitate user interpretation, we introduce the ``Pattern Dice'' as displayed in Fig.~\ref{fig:teaser} \vcircle{b}, which abstracts a group's error pattern into a 3x3 grid of dots, where each row corresponds to one sub-metric of \ref{metric:granularity} and each column from left to right corresponds to a discretized value range. This ingenious summary provides a distilled visual cue, ensuring validators can swiftly identify issues for debugging (\ref{req:r4}).

\item \vcircle{C} \textbf{\viewC{}}: This view provides the 2D distribution of image segments, generated by the segment structuring as discussed in Sec.~\ref{subsec:data_structuring}. As shown in Fig.~\ref{fig:teaser} \vcircle{C}, each point represents one image segment, and the textual semantics of segment groups are displayed on top of each group center, providing straightforward information about the key content of each group (\ref{req:r3}), with a horizontal bar indicating the group's overall segment-label similarity (\ref{metric:seg_label_sim}). Despite addressing \ref{issue:segment} is not the main focus of \systemname{}, in the situation where users conduct in-depth validation, this view also allows users to investigate segment issues by highlighting, checking, and annotating single or groups of segments with issues (\ref{req:r1}, \ref{req:r4}). In response to selections in other views, this view highlights the related segments simultaneously, showing complementary information to assist in the interpretation of data issues (\ref{req:r4}).

\item \vcircle{D} \textbf{\viewD{}}: In concert with \vcircle{C}, this view updates in sync with the other views to show images of corresponding segments, which coordinates with \vcircle{C} for highlighting distributions during users' interaction (Fig.~\ref{fig:teaser} \vcircle{D}). This view allows for in-depth and intuitive exploration, supporting the verification of issues with direct examples (\ref{req:r4}).

\item \vcircle{E} \textbf{\viewE{}}: As shown in Fig.~\ref{fig:teaser} \vcircle{E}, this view provides a semantic summarization of each annotated issue and offers a suite of tools for issue mitigation, including different annotation options and solutions, buttons for adding new options, shortcuts for the annotation history, and options for modifying previous annotations (\ref{req:r4}). It enables validators to seamlessly interact with the dataset, enhancing its quality and ensuring efficient correction of errors. Note that once an issue has been annotated, the corresponding label or group will be displayed in gray as an indication, making it straightforward for human validators to identify unexplored data issues.

\end{enumerate}

\vspace{-10pt}
\subsection{Analytical Workflow}

The design of \systemname{}'s interface consolidates a user-centric analytical workflow for efficacy-driven data issue validation and mitigation (\ref{req:r4}), where each view works coordinately to integrate human expertise in resolving critical label (\ref{issue:label}) and alignment issues (\ref{issue:misalignment}). As illustrated in Fig.~\ref{fig:analytical_workflow}, for addressing issues fall into each category (\ref{issue:label} or \ref{issue:misalignment}), our workflow is structured into three stages: \emph{Explore \& Detect}, \emph{Interpret \& Verify}, and \emph{Annotate \& Solve}, each meticulously crafted to guide users through the data quality enhancement process. In the \emph{Explore \& Detect} stage, validators investigate the label grid or UpSet visualization to identify issues requiring validation. Progressing to the \emph{Interpret \& Verify} stage, following users' selection, our interface displays corresponding distribution overviews or in-depth details, providing the context needed to make informed decisions. Upon verification of an issue, the \emph{Annotate \& Solve} stage provides intuitive interfaces for annotating them and proposing resolutions, ensuring a thorough and effective data refinement. Following these stages, validators can tackle different types of issues systematically, thereby upholding the integrity of the data throughout the validation and mitigation process. Next, we delve into the specifics of how \systemname{} assists validators in resolving label issues and alignment issues, respectively.

\noindent\textbf{Label Issue Validation and Mitigation.} (\ref{issue:label})\label{subsec:interface_label_issue}

\noindent As depicted in the top phase at Fig.~\ref{fig:analytical_workflow}, the workflow starts from the \emph{Explore \& Detect} stage with the \viewA{}, where human validators can brush the histograms to zoom in specific regions, select a potentially wrong cell, and then investigate the label problems grouped by each root.
Upon selection, our visual guidance will lead users to the \emph{Interpret \& Verify} stage, where the \viewC{} and \viewD{} update segment-corresponded points and example images for the verification of flagged issues. Then at the \emph{Annotate \& Solve} stage, solutions can be provided with a few clicks, which will be automatically recorded and presented in the \viewE{}. Finally, when validators confirm the annotated issues and solutions, they are applied to the data set for quality enhancement.

\noindent\textbf{Alignment Issue Validation and Mitigation.} (\ref{issue:misalignment})\label{subsec:interface_misalignment_issue}

\noindent As depicted in the bottom phase at Fig.~\ref{fig:analytical_workflow}, the workflow starts from the \emph{Explore \& Detect} stage with the \viewB{}, where human validators can sort data groups according to specific attribute, explore their patterns, and select one group for detailed inspection. When a group is selected, in the \emph{Interpret \& Verify} stage, \viewC{} and \viewD{} will be updated accordingly, showing semantic descriptions of the selected data group and highlighting their distributions, which support detailed investigation of individual subgroups or segment-label pairs to verify data issues. Similarly, at the \emph{Annotate \& Solve} stage, the verified alignment issues can be annotated and applied following validators' confirmation. 

\vspace{-10pt}
\subsection{Data Issue Resolution}

When human validators confirm and apply their annotations at the end of the analytical workflow, \systemname{} meticulously documents them to apply human-proposed solutions for data issue correction. In our pursuit to improve the quality of data generated by FMs, we consciously refrain from utilizing these models at the error-addressing stage. Instead, we have collaborated with ML experts to devise three principal remedial strategies: (1) removing irrelevant labels, (2) removing problematic segments, and (3) seeking alternative matching segments for misaligned labels. These are provided in our interface as candidate solutions. If humans select from these categories, they are automatically executed to rectify the specified issues. In situations where the solutions fall outside, our specialists step in to review and address them accordingly.

As a prototype framework for addressing this challenging problem, our emphasis is on exploring the efficacy of human intervention and validating its significance in data quality enhancement. This approach allows us to establish a closed-loop system with manageable solutions, underscoring the merit of human contribution. While our current focus is on lightweight interventions, we remain receptive to the adoption of more advanced techniques in future iterations of our framework.

\section{Evaluation}
\label{sec:case_study}
\setlength{\floatsep}{2pt plus 2pt minus 2pt}
\setlength{\textfloatsep}{2pt plus 2pt minus 2pt}
\setlength{\intextsep}{2pt plus 2pt minus 2pt}
\begin{figure}
\setlength{\abovecaptionskip}{0pt}
  \centering\includegraphics[width=\linewidth]{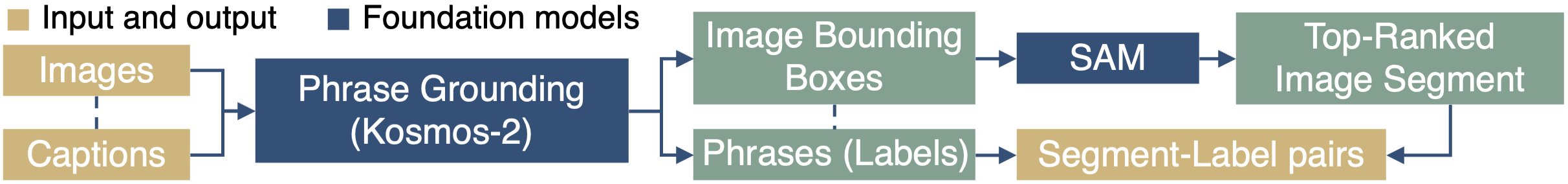}
  \caption{Data pipeline for getting segment-label pairs in our use cases.}
  \label{fig:case_study_data_pipeline}
\end{figure}

In this section, we demonstrate the effectiveness of \systemname{} with two use cases and an expert study. With two datasets in the general and autonomous driving domain, respectively, in use cases, we demonstrate how \systemname{} supports users to conduct data enhancement in different scenarios. Additionally, we invited ten external ML experts who have not seen or used \systemname{} before in our expert study to evaluate our solution.

\begin{enumerate}[align=left, leftmargin=0pt, listparindent=\parindent, labelwidth=0pt, itemindent=0pt, itemsep=0pt, topsep=0pt, label={}, itemsep=1pt, parsep=1pt]

\item \textbf{Data Pipeline.}
In our study, a data generation pipeline was employed to acquire image segment-label data, as depicted in Fig.~\ref{fig:case_study_data_pipeline}. The process begins with the Kosmos-2 performing phrase grounding on paired images and captions, which results in preliminary labels grounded with bounding boxes, prompting SAM to generate corresponding segment masks. This pipeline represents a typical approach for generating segment-label pairs with FMs, serving as the basis for OVIS. With the generated data, we deploy our \systemname{} for two cases.

\end{enumerate}

\vspace{-10pt}
\subsection{Use Case I: Comprehensive Scene Parsing}
Alice is a senior-year PhD student studying machine learning. She is working on an OVIS project with the COCO Stuff dataset~\cite{caesar2018cvpr} and is willing to explore how data quality impacts its performance. Therefore, she conducted a performance-driven analysis with \systemname{}, comparing the OVIS performance based on the original and \systemname{} processed data.

\begin{enumerate}[align=left, leftmargin=0pt, listparindent=\parindent, labelwidth=0pt, itemindent=0pt, itemsep=0pt, topsep=0pt, label={}, itemsep=1pt, parsep=1pt]
\item \textbf{Investigating the validity of labels.}
With the current data pipeline, labels are directly generated by the phase grounding model. Alice does not completely trust the model, doubting if the labels themselves are troublesome. Following the analytical workflow, Alice interacts with the grid visualization of labels in view \vcircle{A} of \systemname{}, as shown in Fig.~\ref{fig:teaser}. She brushed the top histogram to check labels with low segment-label similarity, where she found many cells corresponding to high label frequency, indicating both low-performed and dominant labels. By selecting them and investigating views \vcircle{C} and \vcircle{D}, Alice found many labels are not even nouns and should not be assigned to any segment, such as ``front'', which is assigned to $14,350$ image segments. Somehow many such labels were extracted and paired with segments by the FMs. In addition, Alice noticed a meaningful noun, ``camera'', being spotlighted in a neighbor cell as well. By checking the corresponding segments, she found most of them have no camera included, with content ranging from animals to humans, but they are all ``looking at the camera'' (as shown in Fig.~\ref{fig:teaser}). After exploring neighboring cells, Alice verified $18$ labels with similar issues, annotated them as ``irrelevant labels'', and decided to ``remove label'' (refer to Fig.~\ref{fig:analytical_workflow} for an additional example). By annotating them with a few clicks, Alice moved to other views for further data validation.

\item \textbf{Analyzing segments and label pairing errors.}
Given that the data originated from a singular pipeline rather than multiple resources, Alice speculated that there might be other recurring errors. Observing the view \vcircle{B} \viewB{} (refer to Fig.~\ref{fig:teaser}), she first noticed the three groups of bars, where ``large segment size'' is the highest. She also found the ``error rate'' indicator and thought this was important for error detection. Alice sorted the data by ``error rate'' while grouping by ``large segment size'', and found three groups at the top (refer to Fig.~\ref{fig:teaser}), characterized by ``large segment size'' with ``high error rate''. By investigating further details with views \vcircle{D}, she found ``Group-8'' with ``large segment size'', ``small image-label similarity'', and ``small bounding box-label similarity'' includes segments highlighting background information, where their labels are pretty noisy (refer to Fig.~\ref{fig:teaser}). In addition,
\begin{table}[t]
\setlength{\abovecaptionskip}{0pt}
\setlength{\belowcaptionskip}{-15cm}
\caption{OVIS performance comparison when using the FM-generated data and the \systemname{}-enhanced data.}
\resizebox{\columnwidth}{!}{%
\begin{tabular}{p{0.9cm}<{\centering}p{1.8cm}<{\centering}p{1.8cm}<{\centering}p{1.8cm}<{\centering}p{1.8cm}<{\centering}}
\toprule[0.8pt] 
\multirow{2}{*}{Data} & \multicolumn{2}{c}{Use Case I} & \multicolumn{2}{c}{Use Case II} \\ \cmidrule(l){2-3}  \cmidrule(l){4-5}
 & \textit{mIoU}          & \textit{Pixel Acc}        & \textit{mIoU}      & \textit{Pixel Acc}      \\\toprule[0.5pt]
Original   & 32.39         & 52.01        & 35.55        & 32.63              \\ 
\rowcolor[HTML]{EFEFEF} 
Enhanced & 41.18 (\textcolor{blue}{$\uparrow$ 8.79})   & 62.37 (\textcolor{blue}{$\uparrow$ 10.36})        & 49.90 (\textcolor{blue}{$\uparrow$ 14.35})          & 44.83 (\textcolor{blue}{$\uparrow$ 12.20})                \\
\bottomrule[0.8pt]
\end{tabular}%
}
\label{tab:eval}
\end{table}
upon examining the segment distributions in view \vcircle{C}, she observed that the labels of the segments’ neighbors, which were not highlighted, were accurate. Consequently, she marked this group's issue as ``extremity-sized segmentation'' and the solution as ``find matching labels in neighbors''. Additionally, for ``Group-12'' with ``large segment size'', ``middle image-label similarity'', and ``low bounding box-label similarity'', she found the segments frequently depicted background elements, with labels correctly describing the image but not the segments. For instance, an image of a bus station had its ground segment erroneously labeled as ``a station'', indicating an issue of ``misplaced label''. Alice designated the resolution to ``re-assign label to other segments in image'' for such cases. Continuing this methodical approach, Alice addressed other top-ranked groups, annotating and resolving verified issues.

\item \textbf{Performance enhancement.}
In total, Alice annotated $27$ problematic labels and $20$ misaligned groups. By checking the report generated by \systemname{}, Alice was impressed that errors corresponding to $1.1M$ segment-label pairs have been fixed, which indicates a significant speed-up over manual data curation. After issue correction, Alice conducted the online evaluation(refer to Sec.~\ref{subsec:online_eval}) to compare OVIS performance with \textit{mIoU} and \textit{pixel accuracy}.
As shown in Tab.~\ref{tab:eval}, the model demonstrated improved performance with \systemname{}-enhanced data, confirming the premise that high-quality data is essential for superior model performance.
\end{enumerate}

\begin{figure*}
\setlength{\abovecaptionskip}{0pt}
  \centering\includegraphics[width=\linewidth]{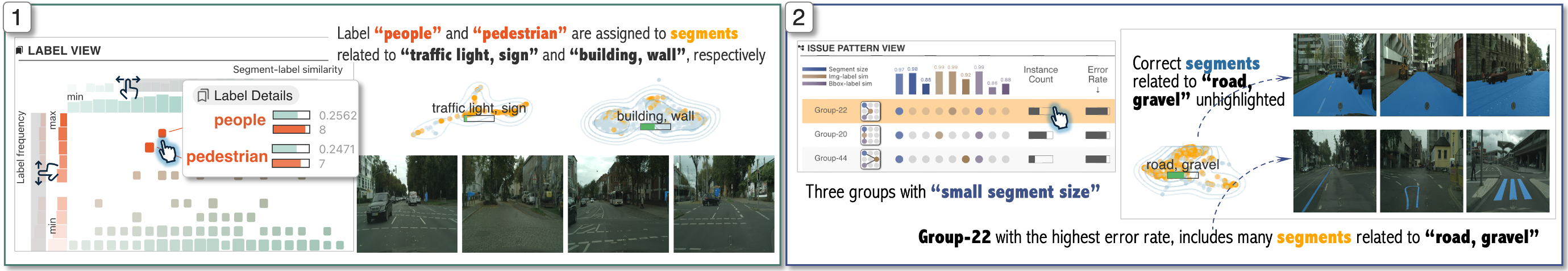}
  \caption{The use case of traffic scene understanding, where the user identified: (1) Labels ``people'', ``pedestrian'' mistakenly assigned to vague image segments related to ``traffic light, sign'' and ``building, wall'', respectively. (2) The problematic subgroup, ``Group-22'', had ``small segment size'' but included many ``road, gravel'' segments, which should be large segments if correctly segmented.}
  \vspace{-10pt}
  \label{fig:case2}
\end{figure*}

\vspace{-10pt}
\subsection{Use Case II: Traffic Scene Understanding}

Kevin is a researcher developing computer vision (CV) methods for autonomous vehicles. He is working on the Cityscapes dataset\cite{cordts2016cityscapes}, which includes traffic scenarios with vehicles, pedestrians, and street signs. Kevin focused on discovering the hidden failures of auto-labeling, particularly those involving critical elements in the driving scene.

\begin{enumerate}[align=left, leftmargin=0pt, listparindent=\parindent, labelwidth=0pt, itemindent=0pt, itemsep=0pt, topsep=0pt, label={}, itemsep=1pt, parsep=1pt]
\item \textbf{Pedestrian identification accuracy.}
Correctly identifying pedestrians is paramount for autonomous driving. As illustrated in Fig.~\ref{fig:case2}, Kevin investigated the \viewA{} to check whether there are problematic labels related to pedestrians. Sifting through the ``segment-label similarity'' histogram, he discovered that ``people'' and ``pedestrian'' lied in two neighboring cells, showing low similarity scores. Diving into the corresponding distributions in the \viewC{}, he found frequent errors for assigning the label ``people'' to ``traffic light, sign'' segments, or assigning the label ``pedestrian'' to ``building, wall'' segments. Curious about the underlying causes, Kevin then checked related samples in the \viewD{} in detail, which led to significant insights---(1) Misalignments involving the label ``people'' often occurred with small, indistinct traffic signs or traffic sign-shaped segments; and (2) For the misalignment between the label ``pedestrian'' and the segments ``building, wall'', there were recurring situations where part of a building or wall is close to pedestrians. A common observation for both is the corresponding segments are vague, due to distance or lighting factors. After verification through the interface, Kevin categorized them as ``vague segments'' and ``misplaced labels'', respectively. Moreover, a review of neighboring cells at the \viewA{} exposed that the label ``person'' was also associated with low ``segment-label similarity'' scores. This led Kevin to conclude that human-related segments were frequently susceptible to mislabeling, warranting a more thorough investigation.

\item \textbf{Other common traffic labeling failures.}
Kevin proceeded to the \viewB{} to check the existence of other segment-label misalignment. He noticed that when ranking the groups decreasingly according to the error rate, many groups corresponding to ``small segment size'' stayed on the top. Initially, he thought this might correspond to the pedestrian identification issue due to the typically small segment size. However, by investigating the \viewC{}, Kevin found a large portion of the segment cluster ``road, gravel'' was highlighted, which was an unexpected pattern, contradicting with the common sense that ``road, gravel'' should have large sizes. Upon examining the cases, Kevin found that the model only captured edges of roads or traffic lines rather than the complete road, yet labeled such edges as ``road''. He decided to visit the \viewC{} to check the entire cluster with semantics ``road'', where he confirmed that for those not highlighted by our framework, the segments were correctly associated with the entire road. He annotated them with ``correct labeling with road'' and ``wrong labeling with road'', respectively. This finding indicated \systemname{}'s accuracy in detecting hidden issues and distinguishing between correct and problematic cases.

\item \textbf{Iterative data enhancement and performance gains.}
Because of the safety concerns, Kevin decided to conduct iterative data enhancement. After mitigating the detected errors, he loaded the updated data back to \systemname{} for the next-round data validation, repeating this process until reaching a satisfactory stage. Lastly, he validated the resulting data with \systemname{} and verified that they had the desired quality, supported by two observations: (1) the overall distribution of ``segment-label similarity'' in the \textit{Label View} was much higher; (2) he had to increase the threshold to $0.8$ to flag data with ``potential issues'', indicating a good alignment between segments and labels in general.
By checking the automatically generated report, he found that \systemname{} helped him correct auto-labeling issues of $15k$ samples. He also conducted the quantitative evaluation to compare training-free OVIS performance with the original and enhanced data as shown in Tab.~\ref{tab:eval}, which confirmed the efficacy of \systemname{} in boosting model performance. He collected the final annotation record and decided to validate other auto-labeling technologies using \systemname{}.
\end{enumerate}

\vspace{-10pt}
\subsection{Expert Study}

As mentioned at the beginning of this section, ten external ML experts were invited to our study, referred to as $E_0 \dots E_9$.
\begin{enumerate}[align=left, leftmargin=0pt, listparindent=\parindent, labelwidth=0pt, itemindent=0pt, itemsep=0pt, topsep=0pt, label={}, itemsep=1pt, parsep=1pt]

\item \textbf{Expert demographics.}
Before the study, We collected detailed demographics of each expert, to contextualize their feedback and ensure that they belong to our target user---ML experts who do not necessarily have visualization/VA knowledge. Among them, $4$ are PhD students and $6$ have obtained PhD degrees. All experts major in Computer Science for their PhD studies. We asked them to self-report their expertise in two areas on a scale of $[0,10]$ ($0$ for ``no knowledge'' and $10$ for ``expert'') and report the statistics as follows:

• Expertise in ML/CV: Md = 9.0, IQR = 1.7;

• Expertise in Visualization/VA: Md = 0.0, IQR = 1.0.\\
For topics they are mostly familiar with, all experts have either used or conducted research on FMs, with other research focuses spanned among: object detection/segmentation ($E_0$-$E_4,E_8$), generative AI ($E_2,E_5$-$E_7,E_9$), medical/biomedical topics ($E_1,E_3$), autonomous driving ($E_0,E_4,E_6$), and 3D reconstruction ($E_1,E_2,E_5$-$E_9$).

\item \textbf{Study setup.}
Each expert was invited to join a one-on-one session, lasting around $60$ minutes. In the session, we started with a $5$-minute introduction of the background, tasks, and interface. Then experts were asked to explore \systemname{} with the public domain case, COCO Stuff, following a think-aloud protocol and exploring as many functionalities as possible. We recorded the study process and noted their comments, paying particular attention to how well the design requirements were met. At the end, we gathered each expert's feedback through a questionnaire with Likert-type questions and an interview.
\end{enumerate}

\setlength{\floatsep}{2pt plus 2pt minus 2pt}
\setlength{\textfloatsep}{2pt plus 2pt minus 2pt}
\setlength{\intextsep}{2pt plus 2pt minus 2pt}
\begin{figure}
\setlength{\abovecaptionskip}{0pt}
  \centering\includegraphics[width=\linewidth]{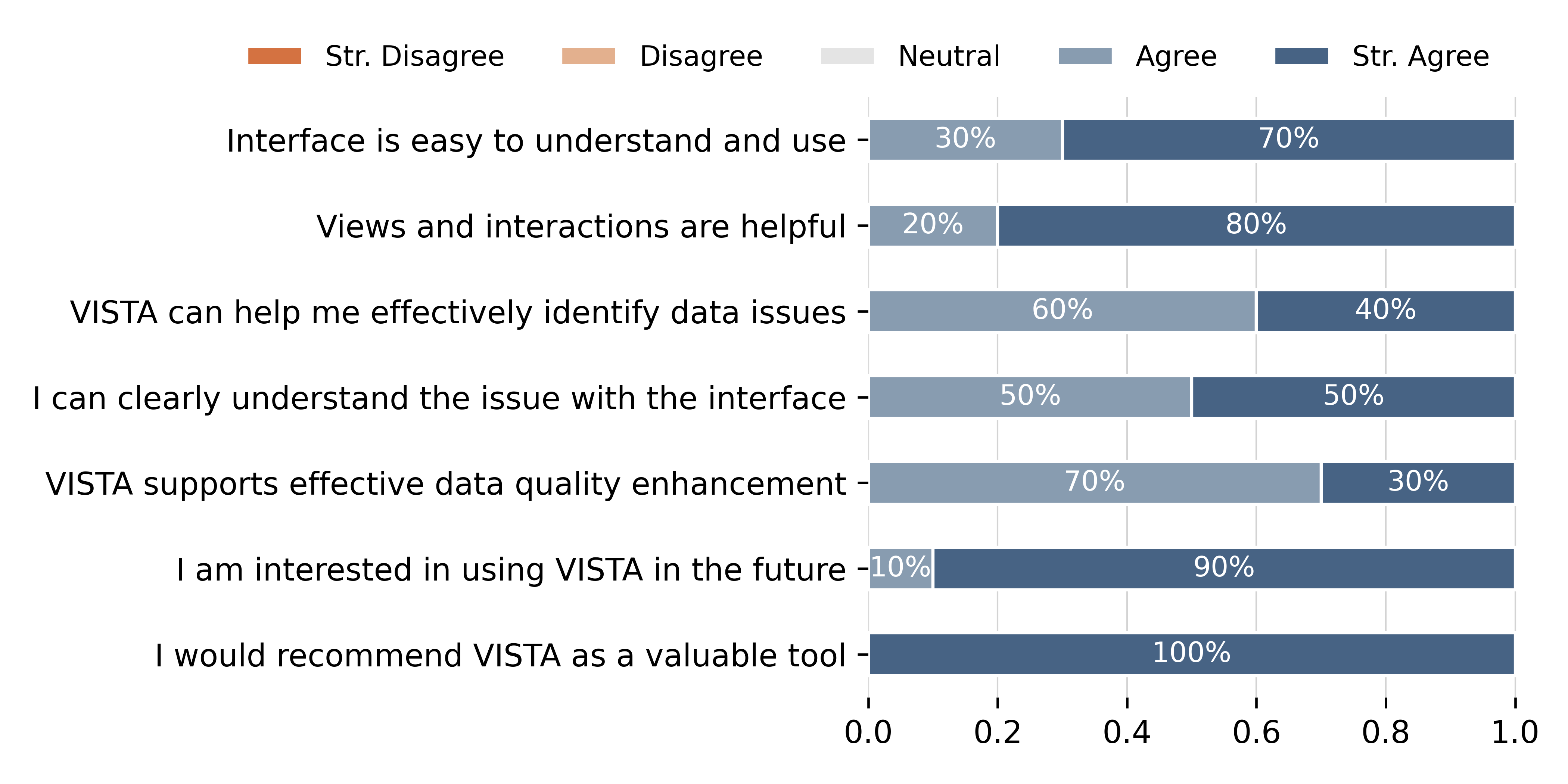}
  \caption{Expert perception of the system according to Likert-type questions.}
  \label{fig:likert}
\end{figure}

Fig.\ref{fig:likert} provides the results of our post-questionnaire, which indicates all experts positively rated the usability and usefulness of \systemname{}. From the interview, we obtained more detailed feedback, and the summarization is provided below.

\begin{enumerate}[align=left, leftmargin=0pt, listparindent=\parindent, labelwidth=0pt, itemindent=0pt, itemsep=0pt, topsep=0pt, label={}, itemsep=1pt, parsep=1pt]

\item \textbf{Detection of hidden errors} (\ref{req:r1}, \ref{req:r2}).
All experts successfully identified a range of data errors with \systemname{}. $E_0$ and $E_1$ commented positively on \systemname{} for its analytical depth in dissecting complex multi-modal data. $E_1$ appreciated the framework’s contribution to improving model robustness, saying \textit{``The key concern of adapting FMs in the medical domain is its uncontrolled nature and you helped mitigate this gap''}. $E_0$ shared the same view---\textit{``The thorough validation from different angles increases my confidence in the data,''} highlighting the importance of multi-perspective data validation as \textit{``crucial but often neglected.''}

\item \textbf{Interpretation and verification of data issues} (\ref{req:r3}).
$E_0$ commented \systemname{} for providing sufficient information to make data issues interpretable. In addition, all experts agreed that our coordinated views facilitated their understanding, with $E_1$ and $E_7$ particularly enjoying the interactive feature. $E_7$ mentioned \textit{``The interface simultaneously showed corresponding information upon my interactions. This greatly helped my task.''} $E9$ talked about the potential of \systemname{} for synthetic data, indicating that our \textit{Embedding} and \textit{Sample Views} can help them validate whether their generated data was realistic.

\item \textbf{Usability of \systemname{}} (\ref{req:r4}).
Following our brief introduction, all experts navigated the system smoothly. Regarding our designed analytical workflow, $E_1$, $E_5$, and $E_8$ described it as \textit{``a clear guidance''}. $E_0$ mentioned the sorting function, \textit{``it allows me to group various features for a closer examination.''} $E_2$ and $E_4$ commended our ``Pattern Dice'' design as \textit{``compact and helpful''}, and $E_4$ said \textit{``it makes identifying patterns super easy.''} We also noticed that although both $E_5$ and $E_{10}$ haven't used any VA system, they promptly finished the data enhancement tasks, proving the usability of \systemname{}.

\item \textbf{Other comments.}
The experts also offered constructive feedback. $E_0$ expressed interest in metric customization. While praising the system’s capability in summarizing prevalent issues, she proposed that \textit{``in later exploring stages, allowing users to define new metrics might uncover trivial problems that are less apparent.''} $E_9$ discussed the possibilities of applying \systemname{} in natural language validation. ``I would be interested to see how to validate LLM-generated data, '' she mentioned. Despite some suggestions might be out-of-scope of this work, we plan to explore these invaluable comments in the future.

\end{enumerate}

\section{Conclusion}
\label{sec:conclusion}
In this work, we introduced \systemname{}, a pioneering visual analytics framework designed to enhance the quality of multi-modal data generated by FMs, particularly focusing on the challenging task of OVIS. 
\systemname{} incorporates the strengths of automated computational processes with human expertise.
Through \systemname{}, we have demonstrated that a carefully designed visual analytics framework can substantially improve the data quality generated by FMs.
Our research underscores the critical necessity of prioritizing data quality over mere data quantity in the era of an unprecedented data explosion.

\section*{Acknowledgments}
This research is sponsored in part by the National Institute of Health under grant no. P41 EB032840 and National Science Foundation via grant no. IIS-2427770.


\bibliographystyle{IEEEtran}
\bibliography{sections/references}

\vspace{-25pt}
\begin{IEEEbiography}[{\includegraphics[width=1in,height=1.25in,clip,keepaspectratio]{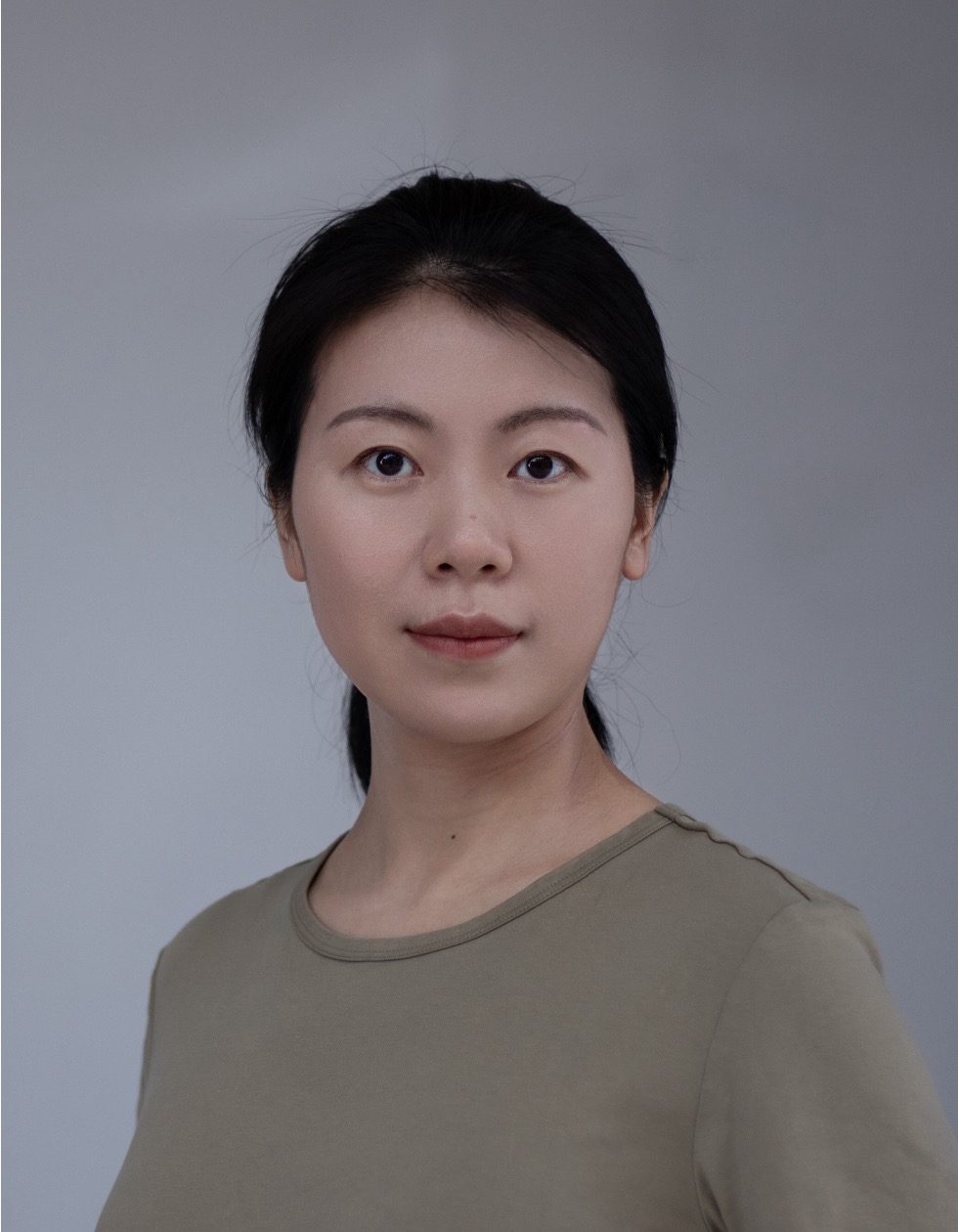}}]%
    {Xiwei Xuan} is a Ph.D. candidate in computer science at the University of California, Davis.
    Before UC Davis, she received her M.S. degree in electrical engineering in 2020 from Washington University in St. Louis.
    Her research spans computer vision, machine learning, and visual analytics, aiming to address the reliability, efficiency, and transparency of machine learning. She is particularly interested in improving data quality and the alignment between human and machine intelligence.
\end{IEEEbiography}
\vspace{-25pt}
\begin{IEEEbiography}[{\includegraphics[width=1in,height=1.25in,clip,keepaspectratio]{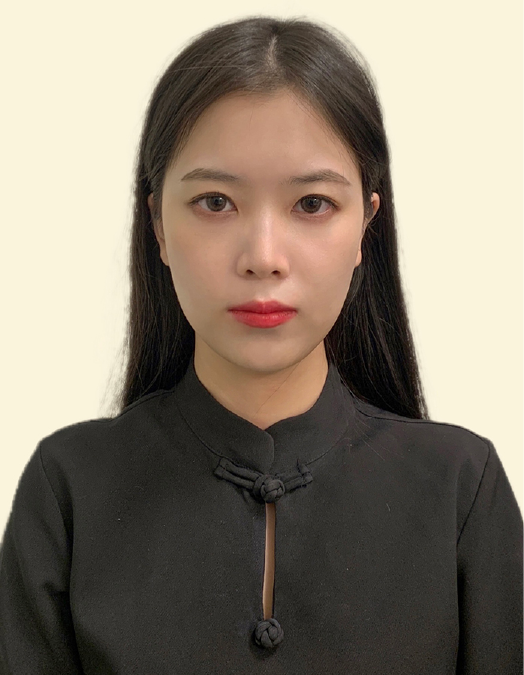}}]%
    {Xiaoqi Wang} is currently a research scientist at Bosch AI center. She received a Bachelor of Science in data analytics from the Ohio State University and a Master of Science in data science from Columbia University. In 2024, she obtained a Ph.D. in computer science at the Ohio State University. Her research interests mainly lie in computer vision, AI explainability and information visualization. Specifically, she is interested in studying image segmentation, GNN explainability and graph visualization. Contact her at wang.5502@osu.edu.
\end{IEEEbiography}
\vspace{-25pt}
\begin{IEEEbiography}[{\includegraphics[width=1in,height=1.25in,clip,keepaspectratio]{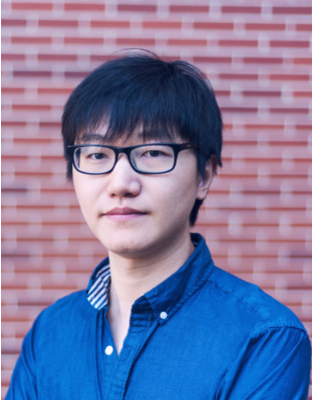}}]%
    {Wenbin He} is a lead research scientist at Bosch. He received his Ph.D. in computer science and engineering from The Ohio State University, advised by Prof. Han-Wei Shen. His research interests include applied artificial intelligence, data analysis and visualization, and computer graphics. Specifically, his work focuses on multi-modal foundation models, visual analytics for explainable AI (XAI), and visual computing with machine learning.
\end{IEEEbiography}
\vspace{-25pt}
\begin{IEEEbiography}[{\includegraphics[width=1in,height=1.25in,clip,keepaspectratio]{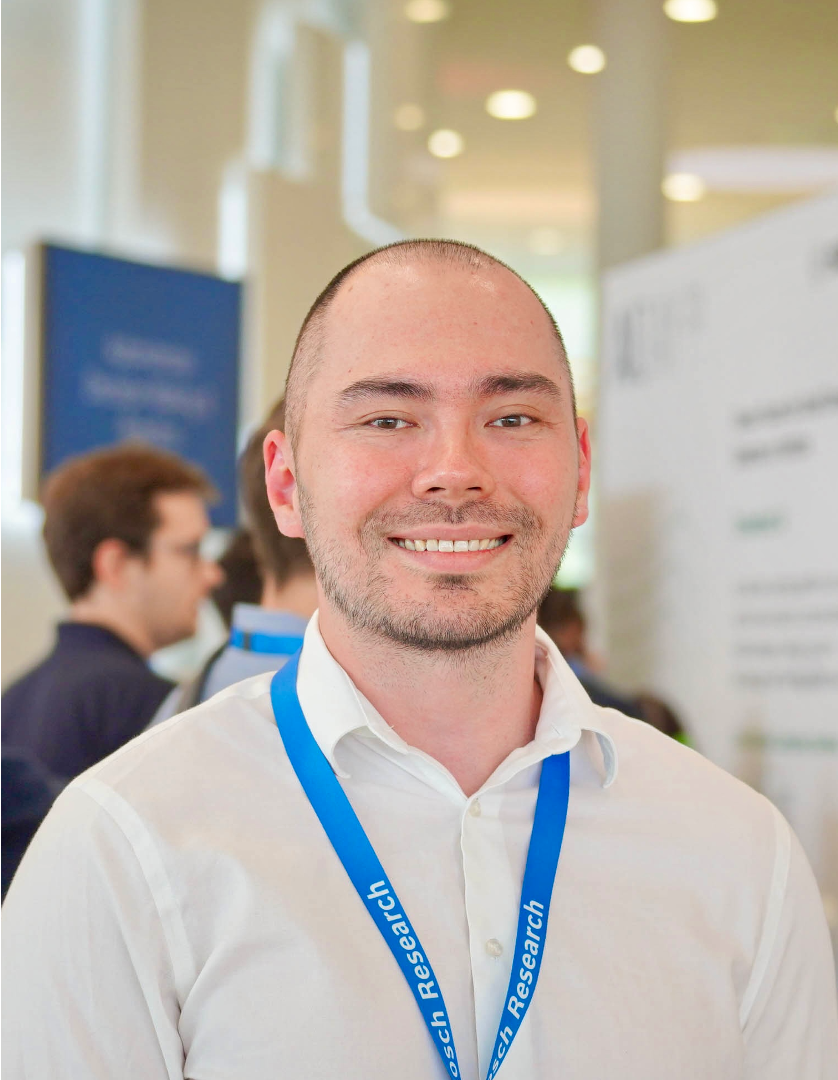}}]%
    {Jorge Piazentin Ono} is a Senior Research Scientist at Bosch. He earned his PhD in Computer Science from New York University and a Master’s degree from the University of Sao Paulo. His research interests include Human-Computer Interaction, Visual Analytics, Explainable AI, and Model Validation. Specifically, his work focuses on leveraging human domain knowledge and insights to improve data quality and model performance.
\end{IEEEbiography}
\vspace{-25pt}
\begin{IEEEbiography}[{\includegraphics[width=1in,height=1.25in,clip,keepaspectratio]{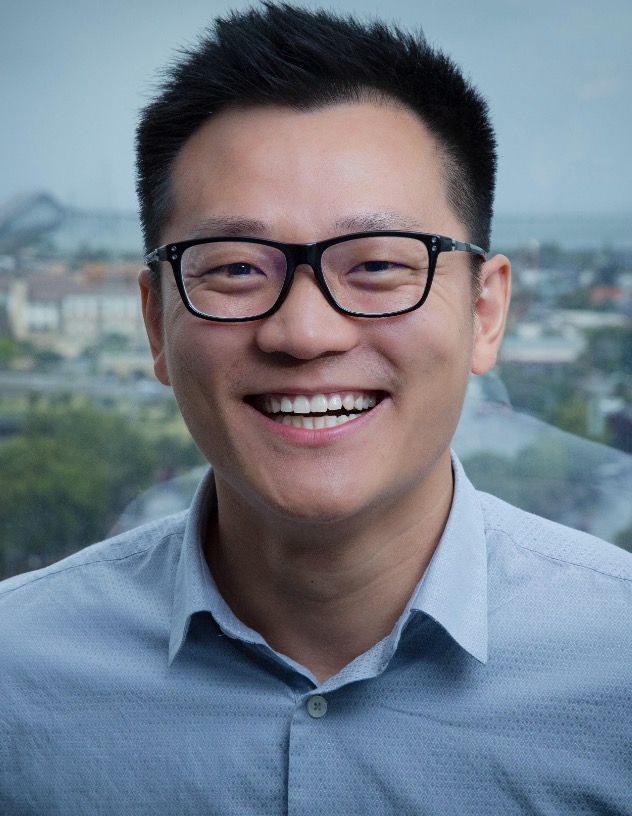}}]%
    {Liang Gou} is the Director of AI at Splunk, a part of Cisco. Previously, he served as a Senior Principal Research Scientist at Bosch, a Principal Research Scientist at Visa Research, and a Research Staff Member at IBM’s Almaden Research Center. Liang holds a Ph.D. in Information Science from Penn State University. His research interests focus on large language models (LLMs), agentic systems, and visual analytics.
\end{IEEEbiography}
\vspace{-25pt}
\begin{IEEEbiography}[{\includegraphics[width=1in,height=1.25in,clip,keepaspectratio]{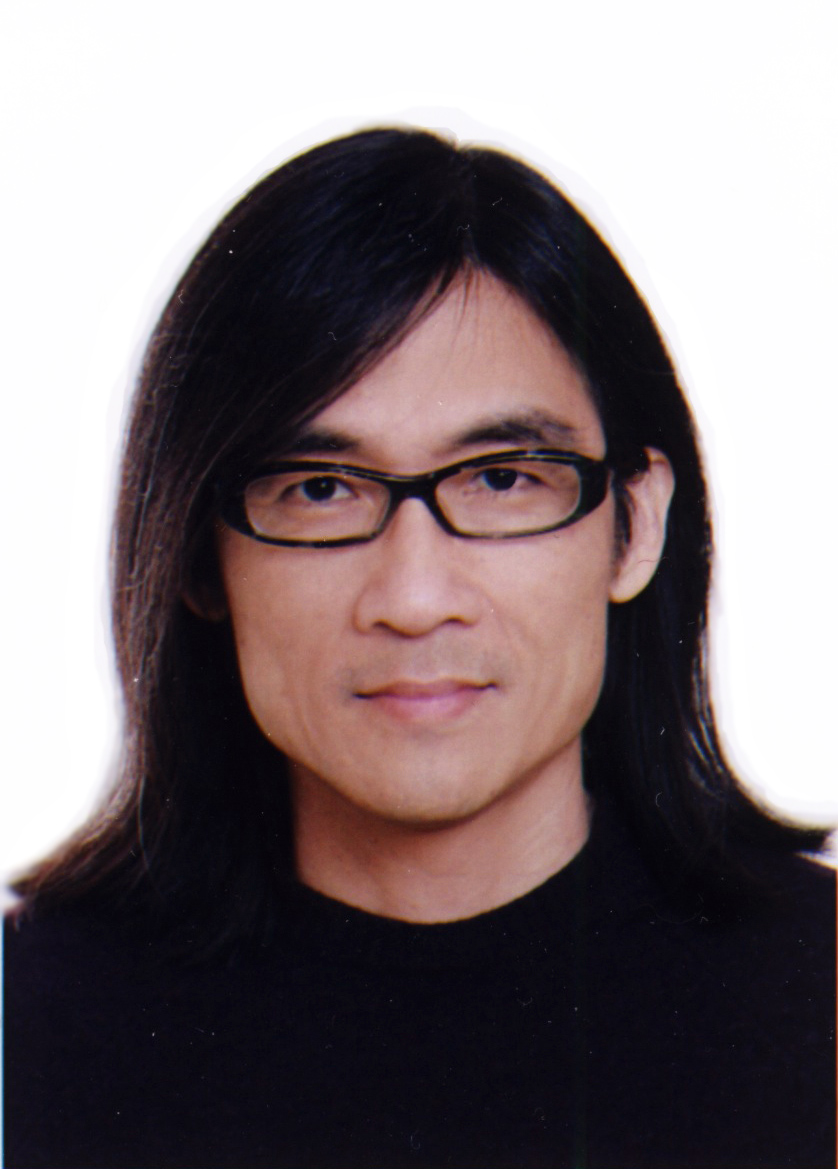}}]%
    {Kwan-Liu Ma} is a distinguished professor of computer science at the University of California, Davis. His research is in the intersection of data visualization, computer graphics, human-computer interaction, and high performance computing. For his significant research accomplishments, Ma received several recognitions including being elected as IEEE Fellow in 2012 and ACM Fellow in 2024, recipient of the IEEE VGTC Visualization Technical Achievement Award in 2013, and inducted to IEEE Visualization Academy in 2019.
\end{IEEEbiography}
\vspace{-25pt}
\begin{IEEEbiography}[{\includegraphics[width=1in,height=1.25in,clip,keepaspectratio]{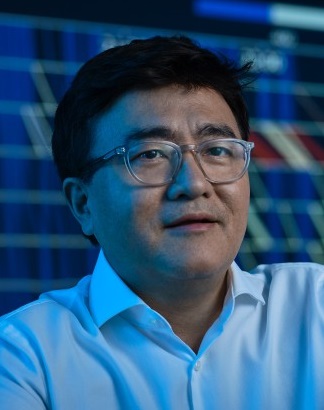}}]%
    {Liu Ren} is currently the Vice President and Chief Scientist of Salable and Assistive AI at Bosch Research North America and Bosch Center for AI (BCAI). He received his Ph.D. in Computer Science from the Computer Science Department at Carnegie Mellon University. His research focuses on AI, computer vision, and visual analytics, among other areas. He has been honored with multiple best paper honorable mention awards (2016,2022) and best paper awards (2018,2020) at IEEE Visualization conferences for his contributions to visual analytics \hspace*{1.15in}\mbox{and XAI}.
\end{IEEEbiography}
\vfill

\end{document}